\documentclass{siamart171218}

\usepackage{lipsum}
\usepackage{amsfonts}
\usepackage{graphicx}
\usepackage{epstopdf}
\usepackage{algorithmic}
\usepackage{subfigure}
\graphicspath{{pics/}{fig/}}

\ifpdf
  \DeclareGraphicsExtensions{.eps,.pdf,.png,.jpg}
\else
  \DeclareGraphicsExtensions{.eps}
\fi

\newsiamremark{remark}{Remark}
\newsiamremark{hypothesis}{Hypothesis}
\crefname{hypothesis}{Hypothesis}{Hypotheses}
\newsiamthm{claim}{Claim}

\title{A scalable deep learning approach for solving high-dimensional dynamic optimal transport\thanks{submitted to the editors April, 2022.
\funding{Chenglong Bao is supported by National Key R\&D Program of China (No. 2021YFA1001300), National Natural Science Foundation of China (Grant No. 11901338), Tsinghua University Initiative Scientific Research Program. Bin Dong are supported in part by National Natural Science Foundation of China (Grant No. 12090022) and Beijing Natural Science Foundation (No. 180001). Zuoqiang Shi is supported by National Natural Science Foundation of China (Grant No. 12071244).}}}

\author{Wei Wan\thanks{School of mathematics and physics, North China Electric Power University.}
\and Yuejin Zhang\thanks{Department of Mathematical Science, Tsinghua University.}
\and Chenglong Bao\thanks{Yau Mathematical Sciences Center, Tsinghua University, and Yaqi Lake Beijing Institute of Mathematical Sciences and Applications.}
\and Bin Dong\thanks{Beijing International Center for Mathematical Research, Peking University, and Center for Machine Learning Research, Peking University.} 
\and Zuoqiang Shi\footnotemark[4]}

\usepackage{amsopn}

\ifpdf
\hypersetup{
  pdftitle={A scalable deep learning approach for solving high-dimensional dynamic optimal transport},
  pdfauthor={mmmm}
}

\begin{document}
\maketitle
\begin{abstract}
The dynamic formulation of optimal transport has attracted growing interests in scientific computing and machine learning, and its computation requires to solve a PDE-constrained optimization problem. The classical Eulerian discretization based approaches 
suffer from the curse of dimensionality, which arises from the approximation of high-dimensional
velocity field. In this work, we propose a deep learning based method to solve the dynamic optimal transport in high dimensional space.
Our method contains three main ingredients: a carefully designed representation of the velocity field, the discretization of the PDE constraint along the characteristics, and the computation of high dimensional integral by Monte Carlo method in each time step. Specifically, in the representation of the velocity field, we apply the classical nodal basis function in time and the deep neural networks in space domain with the $H^1$-norm regularization. This technique promotes the regularity of the velocity field in both time and space such that the discretization along the characteristic remains to be stable during the training process.  Extensive numerical examples have been conducted to test the proposed method. Compared to other solvers of optimal transport, our method could give more accurate results in high dimensional cases and has very good scalability with respect to dimension. Finally, we extend our method to more complicated cases such as crowd motion problem. 
\end{abstract}



\section{Introduction}\label{Introduction} 

Optimal transport (OT) \cite{santambrogio2015optimal,villani2003topics} is an exciting research topic that connects many subjects in mathematics including differential geometry, partial differential equations, optimization, probability theory. Since OT naturally presents a tool to study probability distributions, it has been applied to many tasks in machine learning \cite{torres2021survey}, such as transfer learning \cite{7586038,redko2019optimal}, generative models \cite{2017Wasserstein,salimans2018improving,lei2019geometric}, image processing \cite{papadakis2015optimal} and natural language
processing \cite{fwetdd}.
Despite its numerical success, computing high-dimensional OT remains a challenge. Many 
traditional mesh-based discretization methods \cite{papadakis2014optimal,2021Yu,liu2021multilevel} suffer from the curse of dimensionality. The computation bottleneck limits the applications of OT for various high-dimensional problems, which motivates this work. 

The OT problem is originated from Monge \cite{monge1781memoire} that studies good transportation with minimal cost. Mathematically, let $\mathcal{X}$ and $\mathcal{Y}$ be two complete and seperable metric spaces, $\mu \in \mathcal{P}(\mathcal{X})$ and $\nu \in \mathcal{P}(\mathcal{Y})$ are two probability measures, the \textbf{Monge problem} is to find a transportation map $T:\mathcal{X}\to\mathcal{Y}$ such that it minimizes
\begin{align}\label{MongePrb}
    \inf_T\left\{\int_{\mathcal{X}}c(x,T(x))\mathrm{d}\mu(x): T_\sharp\mu = \nu\right\},
\end{align}
where $c:\mathcal{X}\times\mathcal{Y}\to\mathbb{R}$ is the cost function and $T_\sharp$ is the push forward operator induced by $T$, i.e.\  $\nu(B)=\mu(T^{-1}(B))$ for all measurable set $B\subset\mathcal{Y}$. Directly solving \eqref{MongePrb} is a challenging problem as the \eqref{MongePrb} is highly nonconvex and the minimum does not exist in general.

One special choice of the cost function is the quadratic case, i.e.\ $c(x,y) = \|x-y\|^2$. The seminal work~\cite{brenier1991polar} proves that the optimal map $T$ is the gradient of a convex function that satisfies the Monge-Ampère equation. The Monge-Ampère equation is a fully nonlinear PDE and its numerical solver is highly non-trival. Many numerical methods have been developed based on this approach~\cite{benamou_numerical_2014,prins_least-squares_2015,lindsey_optimal_2017}. However, these method dicretize the computational domain by regular grids which is unaffordable for high dimensional problem. In \cite{gu2016variational}, the variational principle of the OT problem is established by linking OT with power diagram in computational geometry. The resulted variational problem can be solved by convex optimization algorithm with applications to image generation \cite{lei2019geometric,an2019ae,an2020ae}. 


    Instead of finding a transport map, \textbf{Kantorovich formulation}~\cite{2006On} relaxes the Monge problem~\eqref{MongePrb} by introducing a joint distribution $\gamma \in \Pi\left(\mu, \nu\right)$. The coupling set $\Pi\left(\mu, \nu\right)$ is defined as
$$
\Pi\left(\mu,\nu\right)=\left\{\gamma\in \mathcal{P}(\mathcal{X}\times \mathcal{Y}): (\pi_x)_{\#}\gamma =\mu, (\pi_y)_{\#}\gamma = \nu \right\},
$$
where $\pi_x$ and $\pi_y$ are the two projections of $\mathcal{X}\times \mathcal{Y}$ onto $\mathcal{X}$ and $\mathcal{Y}$, respectively. In this case, the Kantorovich formulation is to find $\gamma$ by solving 
\begin{align}\label{KantoPrb}
    \inf _{\gamma} \left\{\int_{\mathcal{X} \times \mathcal{Y}}c(x,y) \mathrm{d}\gamma(x,y):\gamma \in \Pi\left(\mu, \nu\right)\right\}.
\end{align}
It is noted that when $\mu, \nu$ are absolutely continuous with respect to the Lebesgue measure, the Kantorovitch problem \eqref{KantoPrb} has an unique solution \cite{villani2003topics,villani2009optimal} of the form $\pi=(\mathrm{Id} \times T)_{\#} \mu$, where $\mathrm{Id}$ stands for the identity map, and $T$ is the unique minimizer of \eqref{MongePrb}.


When $\mu$ and $\nu$ are discrete distributions, the 
Kantorovitch problem \eqref{KantoPrb} can be formulated as a standard linear program (LP) that can be solved by classical algorithms, e.g.\ simplex method. These methods have relatively high computational cost and do not fully exploited the structure in the Kantorovich formulation \eqref{KantoPrb}. The Sinkhorn method \cite{cuturi2013sinkhorn} introduced the entropy regularization into \eqref{KantoPrb}, that can be efficiently solved by the Sinkhorn algorithm~\cite{1964Sinkhorn}. Sinkhorn method significantly accelerates the computation of OT especially with GPU implementation. As dimension grows, the number of grid points in Sinkhorn method need to increase exponentially to achieve satisfactory accuracy which leads to high computational cost and memory requirement.
For continuous cases, Genevay et al. \cite{genevay2016stochastic} first proposed a stochastic gradient descent to optimize the regularized OT problem by expressing the dual variables in reproducing kernel Hilbert spaces (RKHS). Seguy et al. \cite{seguy2018large} proposed to use deep neural networks to parameterize the dual variables for computing an OT plan and then estimate an optimal map as a neural network learned by approximating the barycentric projection of the obtained OT plan. 
In addition, they theoretically prove the convergence of regularized OT plan 
In recent papers, Makkuva et al. \cite{makkuva2020optimal} proposed a minimax optimization to learn the optimal Kantorovich potential.  
The key idea is to restrict the search over all convex functions with introducing a regularization constraint and approximate the optimal potential by an input convex neural networks (ICNN) \cite{amos2017input}.

The dynamic OT, an alternative formulation of OT, is first proposed in~\cite{benamou2000computational} that is to solve a PDE-constrained optimal control problem. Since the dynamic OT provides a close relationship with fluid mechanics, it has been extended to many applications including unnormalized OT \cite{gangbo2019unnormalized}, mean field games \cite{2007Mean}. Borrowing the theory from continuum mechanics, there are many possibilities for developing OT problems under task-specific constraints, which makes the numerical solver for dynamic OT more attractive. In \cite{benamou2000computational}, it applies an Eulerian discretization and solve the constrained minimization problem with the augmented Lagrangian method. The staggered grid discretization is used in \cite{papadakis2014optimal} that proposed a proximal splitting scheme. Based on~\cite{papadakis2014optimal}, a primal-dual formulation of dynamic OT~\cite{2019Henry} is proposed under the Helmohotz-Hodge decomposition, which enforces the divergence-free constraints during iterations. Recently, Yu et al.~\cite{2021Yu}proposed an accelerated proximal gradient algorithm using multilevel and multigrid strategies to solve dynamic mean field planning problems, which can be viewed as a generalization of OT. 
Above methods can obtain good result in low-dimension cases, but cannot be readily extended to high dimensions.

In recent years, deep learning \cite{schmidhuber2015deep,Goodfellow-et-al-2016} have demonstrated remarkable success on a variety of computational problems ranging from image classification \cite{krizhevsky2012imagenet,sultana2018advancements}, speech recognition \cite{abdel2014convolutional,deng2013new} and natural language processing \cite{collobert2008unified,goldberg2016primer}, to numerical approximations of partial differential equations (PDEs) \cite{weinan2017deep,weinan2018deep,SIRIGNANO20181339,raissi2019physics}. Such numerical results suggest that deep neural networks have powerful approximation ability for high-dimensional functions (see~\cite{SHEN2021160,doi:10.1137/20M134695X,devore2021neural} for the approximation theory).
Based on the link between continuous normalizing flows and dynamic OT, the TrajectoryNet was proposed in \cite{2020TrajectoryNet} and applied for learning cellular dynamics. The network in \cite{2020TrajectoryNet} consists of three fully connected layers with leaky ReLU activations, and the loss function is carefully designed by imposing multiple biological priors. The regularized neural ordinary differential equation (RNODE) is proposed in \cite{finlayHowTrainYour2020} by penalizing the Frobenius norm of the Jacobian of the velocity field, which improves the training stability. The Frobenius norm and the divergence of the velocity field is calculated by Monte Carlo trace estimator. Very recently, Ruthotto et al. \cite{Ruthotto9183} proposed a machine learning based method for solving high-dimensional mean field game and mean field control problems (known as MFGnet).
In their work, a neural network is proposed to represent the Kantorovich potential whose gradient is the velocity field and the Hamilton-Jacobi-Bellman (HJB) equation is employed as a regularization. Numerically, it needs to compute the second derivative of the potential model and the number of training samples and the running time per iteration have noticeable growth as the dimension increases. 

In this work, we propose a new deep learning approach for solving the high-dimensional dynamic OT. We adopt the Lagrangian discretization to solve the continuity equation and use Monte Carlo sampling to approximate the integrals in high dimensions rather than mesh-based discretization. Moreover we carefully designed a neural network for parametrizing the velocity field. Our approach only needs to calculate the first derivative of the neural network with respect to the input. In summary, the contributions of this paper are as follow.
\begin{itemize}
    \item We present a novel approach to parameterize the velocity field by
    combining neural network in space and linear finite element basis in time. Based on the priori information that the characteristics of the optimal solution are straight, we introduce the Frobenius norm of the Jacobian of the velocity field as the regularization to improve generalization.
    \item Our proposed framework is easy to implement and can be extended to more complicated problem, such as crowd motion. The proposed algorithm has much lower computational cost while the accuracy is comparable with state-of-the-art methods, Sinkhorn and MFGnet, especially for high-dimensional cases.
\end{itemize}

The rest of our paper is organized as follows. In Section \ref{TheProposedMethod}, we introduce the dynamic OT problem, the proposed model parameterized by a neural network, and the back propagation method for solving the dynamic OT problems. Experimental results and some details about implementation are summarized in Section \ref{Experimentresult}. Finally, we conclude this work in Section \ref{ConclusionandFutureWork}.

\section{The Proposed Framework to Solve Dynamic OT}\label{TheProposedMethod}

\subsection{Preliminaries on Dynamic OT}\label{PDOT}

In this paper, we mainly consider the dynamic OT problem introduced by Benamou and Brenier \cite{benamou2000computational}. The basic problem can be formulated as
follows:
Given two probability densities $\rho_{0}$ and $\rho_{1}$, we want to find the density $\rho: \mathbb{R}^{d} \times [0,1] \rightarrow \mathbb{R} $ and the velocity field $\boldsymbol{v}:\mathbb{R}^{d}\times [0,1]  \rightarrow \mathbb{R}^{d}$ that transport
the mass of $\rho_{0}$ to the mass of $\rho_{1}$ at minimal transport cost $\mathcal{E}(\rho,\boldsymbol{v})$, i.e., 
\begin{align}\label{dOT}
\min_{(\rho, \boldsymbol{v}) \in \mathcal{C}(\rho_0,\rho_1)} \mathcal{E}(\rho,\boldsymbol{v}):= \int_0^1 \int_{\Omega} |\boldsymbol{v}(x,t)|^2\rho(x,t)\mathrm{d}x \mathrm{d}t
\end{align}
and any pair of $(\rho, \boldsymbol{v}) \in \mathcal{C}\left(\rho_{0}, \rho_{1}\right)$ satisfies continuity equation with initial and terminal densities of $\rho$ being $\rho_{0}, \rho_{1}$ provided as:
\begin{align}\label{massconserv}
    \mathcal{C}(\rho_0,\rho_1):=\{(\rho, \boldsymbol{v}): \partial_t \rho+\nabla\cdot (\rho\boldsymbol{v})=0, \rho(\cdot,0)=\rho_0,\rho(\cdot,1)=\rho_1\}.
\end{align}
Note that in the whole paper, $\nabla$ and $\nabla\cdot$ denote the gradient and the divergence w.r.t. the space variable $x$, respectively.
In addition, the relationship of the optimal map $T$ and the optimal solution $(\rho,\boldsymbol{v})$ is that:
\begin{align}
    \rho(\cdot,t)=(T_{t})_{\#}\rho_{0} ~\text{with}~ T_{t}(x)=(1-t)x+tT(x), \label{opTt}\\
    v_{t}=\left(\frac{d}{d t} T_{t}\right) \circ T_{t}^{-1}=(T-\mathrm{id}) \circ T_{t}^{-1}.
\end{align}

Note that $\mathcal{E}(\rho,\boldsymbol{v})$ denotes the generalized kinetic energy and the optimal $\mathcal{E}(\rho,\boldsymbol{v})$ gives the square of the $L^2$ Wasserstein distance.
This fluid mechanics formulation provides a natural time interpolation which is very useful in many applications.

\subsection{Relaxation of Dynamic OT}\label{Se:RoDOT}
First, we relax the constraint on the terminal density into an implicit condition by introducing a Kullback–Leibler (KL) divergence penalty term:
\begin{align*}
\mathcal{P}(\rho)&:= \lambda \int_{\Omega}\rho(x,1)\log \frac{\rho(x,1)}{\rho_1(x)}
\mathrm{d}x.
\end{align*}
Note that $\mathcal{P}(\rho)$ is the KL divergence between $\rho(\cdot,1)$ and $\rho_{1}$ up to a sufficiently large positive constant $\lambda$, which is the penalty parameter.

From \eqref{opTt} we know that the characteristic of the optimal solution $T_{t}$ is a straight line connecting the source point $x$ and  the target point $T(x)$.
However, the integrated kinetic energy $\mathcal{E}(\rho,\boldsymbol{v})$ forces the dynamics to travel in straight lines only on the training data. In order to improve generalization during training and enhance the smoothness of the velocity field, we introduce the Frobenius norm of the Jacobian of the velocity field as the regularization $\mathcal{R} (\boldsymbol{v})$, i.e.,
\begin{align*}
    \mathcal{R} (\boldsymbol{v}):= \alpha \int_{0}^{1}\int_{\Omega} |\nabla \boldsymbol{v}(x,t)|_{F}^{2} \mathrm{d}x
    \mathrm{d}t,
\end{align*}
where $|*|_{F}$ denotes Frobenius norm and $\alpha$ is a positive regularization parameter.

By relaxing the terminal density as an implicit condition penalized by KL divergence term $\mathcal{P}(\rho)$ and introducing the regularization term $\mathcal{R} (\boldsymbol{v})$, we can get the following optimal control formulation:
\begin{align}\label{objective}
\min_{(\boldsymbol{v},\rho) \in \mathcal{C}(\rho_0)}\mathcal{F}(\rho,\boldsymbol{v}) = \mathcal{E}(\rho, \boldsymbol{v}) + \mathcal{P}(\rho) + \mathcal{R}(\boldsymbol{v}),
\end{align}
where $\mathcal{C}(\rho_0):=\{(\rho, \boldsymbol{v}): \partial_t \rho+\nabla\cdot (\rho \boldsymbol{v})=0, \rho(\cdot,0)=\rho_0\}$.

\subsection{Lagrangian Discretization}
In order to solve the high-dimensional OT problems, we naturally adopt the Lagrangian discretization method instead of the traditional mesh-based method. Because the Lagrangian discretization is a sampling method that can reduce memory usage by overcoming the spatial discretization into grids.
Firstly, the continuity equation in $\mathcal{C}(\rho_0)$ can be solved along the characteristics:
\begin{align}
&\frac{\mathrm{d}}{\mathrm{d}t}\mathbf{z}(x,t)=\boldsymbol{v}(\mathbf{z}(x,t),t),\mathbf{z}(x,0)=x,  \label{char}\\
&\frac{\mathrm{d}}{\mathrm{d}t}\rho(\mathbf{z}(x,t),t)=-\rho(\mathbf{z}(x,t),t)\nabla\cdot \boldsymbol{v}(\mathbf{z}(x,t),t),~\rho(\mathbf{z}(x,0),0)=\rho_0(x). \label{rhos}
\end{align}
Note that the equation  \eqref{rhos} is equivalent to
\begin{align}
\frac{\mathrm{d}}{\mathrm{d}t}\ln\rho(\mathbf{z}(x,t),t)=-\nabla\cdot \boldsymbol{v}(\mathbf{z}(x,t),t),~\ln\rho(\mathbf{z}(x,0),0)=\ln\rho_0(x). \label{lnrhos}
\end{align}
In fact, when the dimension is very high, since $\rho_0$ is a positive number very close to 0, there will be a numerical overflow problem when solving equation \eqref{rhos}, so what we actually solve is equation \eqref{lnrhos}.

Together with the Lagrangian method \eqref{char} and \eqref{rhos}, this leads to an optimization problem with respect to the velocity field $\boldsymbol{v}$, which we will model with a neural network.
In addition, we use the Monte Carlo method for integration and suppose $x_1,x_2,...,x_r$ are samples drawn from $\rho_0$. Along the characteristics, the density satisfies 
\begin{align}\label{var_subst}
\rho(\mathbf{z}(x,t),t)\det(\nabla \mathbf{z}(x,t))= \rho_0(x)
\end{align}
for all $t \in [0,1]$. Thus,
\begin{align*}\label{disE}
 \mathcal{E}(\rho,\boldsymbol{v})&= \int_0^1 \int_{\Omega} |\boldsymbol{v}(x,t)|^2\rho(x,t)\mathrm{d}x \mathrm{d}t\\
 &=\int_0^1 \int_{\Omega} |\boldsymbol{v}(\mathbf{z}(x,t),t)|^2\rho(\mathbf{z}(x,t),t)\det(\nabla \mathbf{z}(x,t))\mathrm{d}x \mathrm{d}t\\
 &=\int_0^1 \int_{\Omega} |\boldsymbol{v}(\mathbf{z}(x,t),t)|^2\rho_0(x)\mathrm{d}x \mathrm{d}t\\
 & = \int_0^1 \mathbb{E}_{\mathbf{x} \sim \rho_0}\big[ |\boldsymbol{v}(\mathbf{z}(\mathbf{x},t),t)|^2 \big] \mathrm{d}t\\
 & \approx \int_0^1 \sum_{i=1}^r \frac{1 }{r} |\boldsymbol{v}(\mathbf{z}(x_i,t),t)|^2 \mathrm{d}t.
\end{align*}
Similarly, we can get
\begin{align*}
\mathcal{P}(\rho)& = \lambda \int_{\Omega}\rho(x,1)\log \frac{\rho(x,1)}{\rho_1(x)}
\mathrm{d}x\\
&=\lambda \int_{\Omega}\log \frac{\rho(\mathbf{z}(x,1))}{\rho_1(\mathbf{z}(x,1))}\rho(\mathbf{z}(x,1),1) \det(\nabla \mathbf{z}(x,1))
\mathrm{d}x\\
&=\lambda \int_{\Omega}\log \frac{\rho(\mathbf{z}(x,1))}{\rho_1(\mathbf{z}(x,1))}\rho_0(x)
\mathrm{d}x\\
&=\lambda \mathbb{E}_{\mathbf{x}\sim \rho_0}\big[\log \frac{\rho(\mathbf{z}(\mathbf{x},1)}{\rho_1(\mathbf{z}(\mathbf{x},1)}\big]\\
& \approx \lambda \sum_{i=1}^r \frac{1}{r} \log \frac{\rho(\mathbf{z}(x_i,1),1)}{\rho_1(\mathbf{z}(x_i,1))}.
\end{align*}
Suppose that $y_1,y_2,..,y_s$ are samples drawn from a uniform distribution whose density is $P_0(y)=P_0$, which is a constant on the support set of the uniform distribution.
\begin{align*}
\mathcal{R} (\boldsymbol{v})& = \alpha \int_{0}^{1}\int_{\Omega} |\nabla \boldsymbol{v}(x,t)|_{F}^{2} \mathrm{d}x \mathrm{d}t\\
&=\alpha \int_{0}^{1} \mathbb{E}_{\mathbf{y}\sim P_0}\big[\frac{|\nabla \boldsymbol{v}(\mathbf{y},t)|_{F}^{2}}{P_0(\mathbf{y})}  \big] \mathrm{d}t\\
& \approx \frac{\alpha}{s P_0}\int_0^1\sum_{j=1}^s|\nabla \boldsymbol{v}(y_j,t)|_F^2 \mathrm{d}t.
\end{align*}
Accordingly, we get the following semi-discretization version of OT problem:
\begin{align}\label{prov} 
 \min_{\boldsymbol{v} }\overline{\mathcal{F}} (\boldsymbol{v},\rho) :=  \overline{\mathcal{E}} (\boldsymbol{v},\rho)  + 
 \overline{\mathcal{P}} (\rho) + \overline{\mathcal{R}} (\boldsymbol{v}),
\end{align} 
where $\overline{\mathcal{E}} (\boldsymbol{v},\rho),  \overline{\mathcal{P}} (\rho)$ and $\overline{\mathcal{R}} (\boldsymbol{v})$ are the discretization versions of $\mathcal{E}(\boldsymbol{v},\rho), \mathcal{P}(\rho)$ and $\mathcal{R}(\boldsymbol{v})$ in the space direction respectively, i.e.,
\begin{align}
\overline{\mathcal{E}} (\boldsymbol{v},\rho)&:= \int_0^1 \sum_{i=1}^r \frac{1}{r}|\boldsymbol{v}(\mathbf{z}(x_i,t),t)|^2 \mathrm{d}t, \label{semdisE}\\
\overline{\mathcal{P}} (\rho)&:= {\lambda}\sum_{i=1}^r \frac{1}{r} \log \frac{\rho(\mathbf{z}(x_i,1),1)}{\rho_1(\mathbf{z}(x_i,1))},\label{semdisP}\\
\overline{\mathcal{R}} (\boldsymbol{v})&:= \frac{\alpha}{s P_0}\int_0^1\sum_{j=1}^s|\nabla \boldsymbol{v}(y_j,t)|_F^2 \mathrm{d}t, 
\end{align}
where $\rho(\mathbf{z}(x_i,t),t)$ can be obtained from the solution of \eqref{lnrhos} along the characteristics $\mathbf{z}(x_i,t)$. Then, \eqref{char} and \eqref{lnrhos} can be reformulated as the following constraints:
\begin{align}
 &\frac{\mathrm{d}}{\mathrm{d}t}\mathbf{z}(x_i,t)=\boldsymbol{v}(\mathbf{z}(x_i,t),t),~\mathbf{z}(x_i,0)=x_i, ~x_i \sim \rho_0(x_{i}), \label{forwd1}\\
 &\frac{\mathrm{d}}{\mathrm{d}t}\ln\rho(\mathbf{z}(x_i,t),t)=-\nabla \cdot \boldsymbol{v}(\mathbf{z}(x_i,t),t),~\ln\rho(\mathbf{z}(x_i,0),0)=\ln\rho_0(x_i).\label{forwd2}
 \end{align}
\subsubsection{Importance Sampling}
When the distribution $\rho_0$ is complex and difficult to sample, or when we want to use the information of the whole region, we can sample from another distribution $\mu_0$, such as uniform distribution. At this time, we need to solve one more ODE
\begin{align*}
     &\frac{\mathrm{d}}{\mathrm{d}t}\ln\mu(\mathbf{z}(x_i,t),t)=-\nabla \cdot \boldsymbol{v}(\mathbf{z}(x_i,t),t),~\ln\mu(\mathbf{z}(x_i,0),0)=\ln\mu_0(x_i).\label{forwd3}
\end{align*}
Accordingly, the objective functions are modified to
\begin{align*}
\overline{\mathcal{E}} (\boldsymbol{v},\rho)&= \int_0^1 \sum_{i=1}^r w_i(t)|\boldsymbol{v}(\mathbf{z}(x_i,t),t)|^2 \mathrm{d}t,\\
\overline{\mathcal{P}} (\rho)&= {\lambda}\sum_{i=1}^r w_i(1) \log \frac{\rho(\mathbf{z}(x_i,1),1)}{\rho_1(\mathbf{z}(x_i,1))},
\end{align*}
where
\begin{align*}
    w_i(t) = \frac{\rho(\mathbf{z}(x_i,t),t)}{\mu(\mathbf{z}(x_i,t),t)r}.
\end{align*}

 \subsection{Gradient Calculation}
In this paper, we parameterize the velocity field $\boldsymbol{v}$ by using a neural network $\boldsymbol{v}(x,t;\theta)$, where $\theta$ denotes the trainable parameters. The structure of the neural network will be discussed in the next section. Then, the original optimization problem for the velocity field $\boldsymbol{v}$ \eqref{prov} can be changed into a deep learning problem. Next, in order to calculate the gradient of the loss function with respect to $\theta$, we have adopted two different methods: the back propagation method and the adjoint state method, which correspond to the discrete-then-optimize and optimize-then-discrete \cite{2019ANODE,Onken2020DO}, respectively.
Gholami et al. \cite{2019ANODE} and Onken et al. \cite{Onken2020DO} give a thorough discussion of the difference between the discretize-then-optimize and optimize-then-discretize approaches and suggest that the discretize-then-optimize approach is preferable due to the guaranteed accuracy of gradients.
In Section \ref{AppAdSM} and \ref{AppCwBP}, we introduce the adjoint state method and compare it with the back propagation method in numerical experiments for the Gaussian problem. 
Our experiments show that the back propagation method can achieve superior results to the adjoint state method at reduced computational costs, which aligns with the results achieved in \cite{2019ANODE} and \cite{Onken2020DO}.

 \subsubsection{Back Propagation}
 In the discrete-then-optimize method, back propagation is realized by using automatic differentiation, which traverses the computational graph backward in time and is used commonly in machine learning frameworks. Actually, the discretization of the forward propagation completely determines
this process.

Since \eqref{prov} is the semi-discretization objective function, we can simplify it as 
\begin{align}
\overline{\mathcal{F}} (\theta) = \int_0^1 f(t;\theta) \mathrm{d}t + 
{\lambda}\sum_{i=1}^r w_i(1) \log \frac{\rho(\mathbf{z}(x_i,1),1)}{\rho_1(\mathbf{z}(x_i,1))},
\label{semdis2}
\end{align}
where
\begin{align*}
f(t;\theta) &= \sum_{i=1}^r w_i(t)|\boldsymbol{v}(\mathbf{z}(x_i,t),t;\theta)|^2 + \frac{\alpha}{s P_0}\sum_{j=1}^s|\nabla \boldsymbol{v}(y_j,t;\theta)|_F^2.
\end{align*}
We use the composite Simpson formula to discrete the semi-discretization objective function \eqref{semdis2}.
\begin{align*}
\int_{0}^{1} f(t) \mathrm{d}t & = \sum_{n=0}^{N-1} \int_{t_n}^{t_{n+1}} f(t) \mathrm{d}t
\approx \frac{h}{6}\sum_{n=0}^{N-1}[ f(t_{n})+4f(t_{n}+\frac{h}{2})+ f(t_{n+1})],
\end{align*}
where $h = \frac{1}{N}, t_{n} = nh, n = 0,\dotsc,N$.

\begin{figure}[h]\label{figBP}
\begin{center}
\includegraphics[width=12cm]{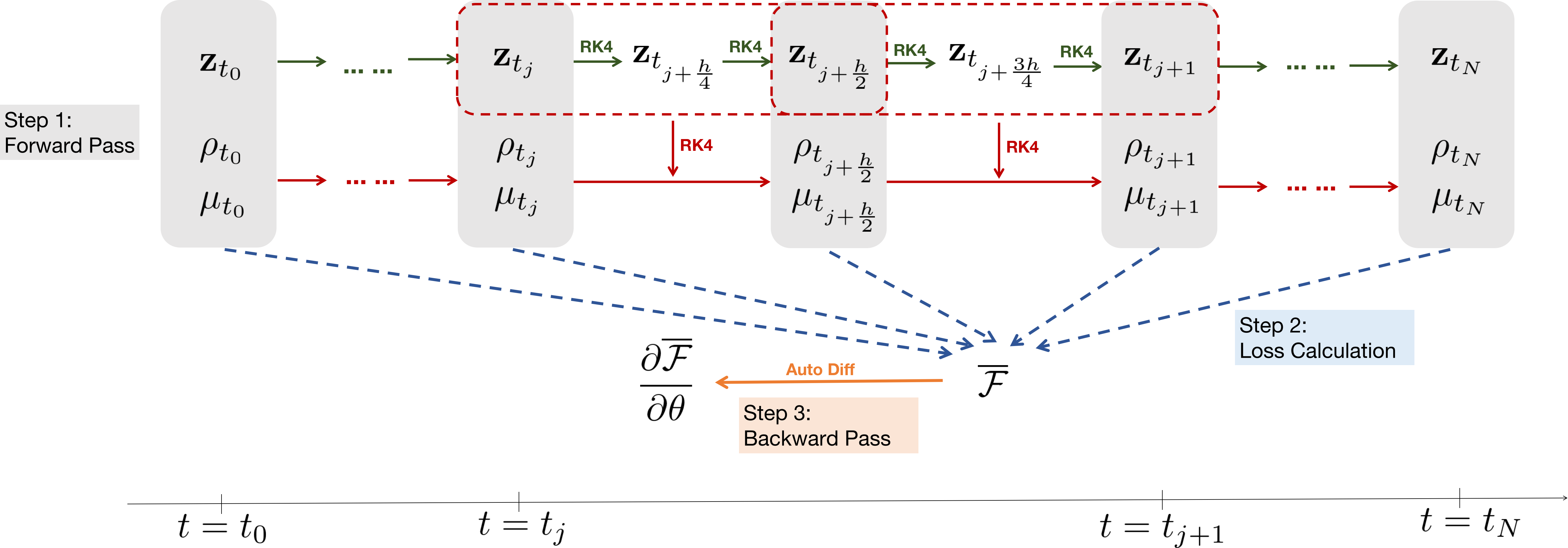}
\caption{The illustration of the back propagation algorithm.}
\end{center}
\end{figure}

Next, we perform the fourth-order explicit Runge-Kutta scheme (RK4) to solve ODE constrains \eqref{forwd1} and \eqref{forwd2}. Then, we can further adopt the Simpson formula to calculate the loss function $\overline{\mathcal{F}}(\theta)$ from \eqref{semdis2}. Finally, the automatic differentiation technique can be used to compute the gradients with respect to the network parameters $\frac{\partial \overline{\mathcal{F}}}{\partial \theta}$, which will be used to update $\theta$. The computational flow chart for calculating the gradients using the back propagation method is shown in Figure \ref{figBP}.
Then, the parameters $\theta$ in the neural network can be optimized by some popular optimizor, such as SGD, Adam, BFGS and so on.

\subsubsection{Adjoint State Method}
\label{AppAdSM}

The discrete-then-optimizer method is common in neural networks and the back propagation is easy to implement, especially when automatic differentiation can be used. However, the back propagation can be memory inefficient because of the potentially huge computation graph. In order to efficiently calculate the gradient $\frac{\partial \overline{\mathcal{F}}}{\partial \theta}$, the optimizer-then-discrete approach, i.e. the adjoint state method has also been introduced in this section. In the adjoint state method, the gradients are computed by numerically solving the adjoint equation. In all of the experiment in this section, we sample $x_{i}$ drawn from $\rho_{0}$, thus $\omega_{i}(t)=\frac{1}{r},~ \forall t\in[0,1], ~i=1,\cdots,r$.
We can show the gradient calculation in the following proposition.
\begin{proposition}\label{adjoint}
		If $\mu_{0}=\rho_{0}$, the gradient of $\overline{\mathcal{F}} (\theta)$ is calculated by the formula
\begin{align}\label{adjoint_grad}
	\frac{\partial \overline{\mathcal{F}} (\theta)}{\partial
	\theta} = &
	\frac{1}{r}\sum_{i=1}^{r} \int_0^1 [(2\boldsymbol{v}(\mathbf{z}_i(t),t;\theta)-U_i(t))^{T}\boldsymbol{v}_{\theta}(\mathbf{z}_i(t),t;\theta)-\lambda \nabla \cdot \boldsymbol{v}_{\theta}(\mathbf{z}_i(t),t;\theta)] \mathrm{d}t \notag\\
	&+\frac{\alpha}{sP_{0}}\sum_{j=1}^{s}\int_{0}^{1}2\left \langle \nabla\boldsymbol{v}_{\theta}(y_{j},t;\theta),\nabla \boldsymbol{v}(y_{j},t)  \right \rangle  \mathrm{d} t.
\end{align}
Here we denote $\mathbf{z}(x_i,t)$ by $\mathbf{z}_i(t)$ for simplification. Moreover,
$\left \langle \cdot,\cdot  \right \rangle $ is defined as
\begin{align*}
    \left \langle A,B  \right \rangle = \sum_{i,j}A_{i,j}B_{i,j}
\end{align*}
and $A,B$ are two matrices with the same size.
$U_i(t)$ is the solution of the adjoint equation
\begin{align}\label{adjoint_eq}
\left\{
\begin{aligned}
\dfrac{\mathrm{d}U_i(t)}{\mathrm{d}t}&+\nabla^T\boldsymbol{v}(\mathbf{z}_i(t),t)U_i(t)=[2(\nabla \boldsymbol{v})^{T} \boldsymbol{v}-\lambda\nabla(\nabla \cdot \boldsymbol{v})](\mathbf{z}_i(t),t;\theta), \\
U(1)&= \lambda\frac{\nabla \rho_1(\mathbf{z}_i(1))}{\rho_1(\mathbf{z}_i(1))}.
\end{aligned}
\right.
\end{align}
\end{proposition}

\begin{figure}[h]
\begin{center}
\includegraphics[width=12cm]{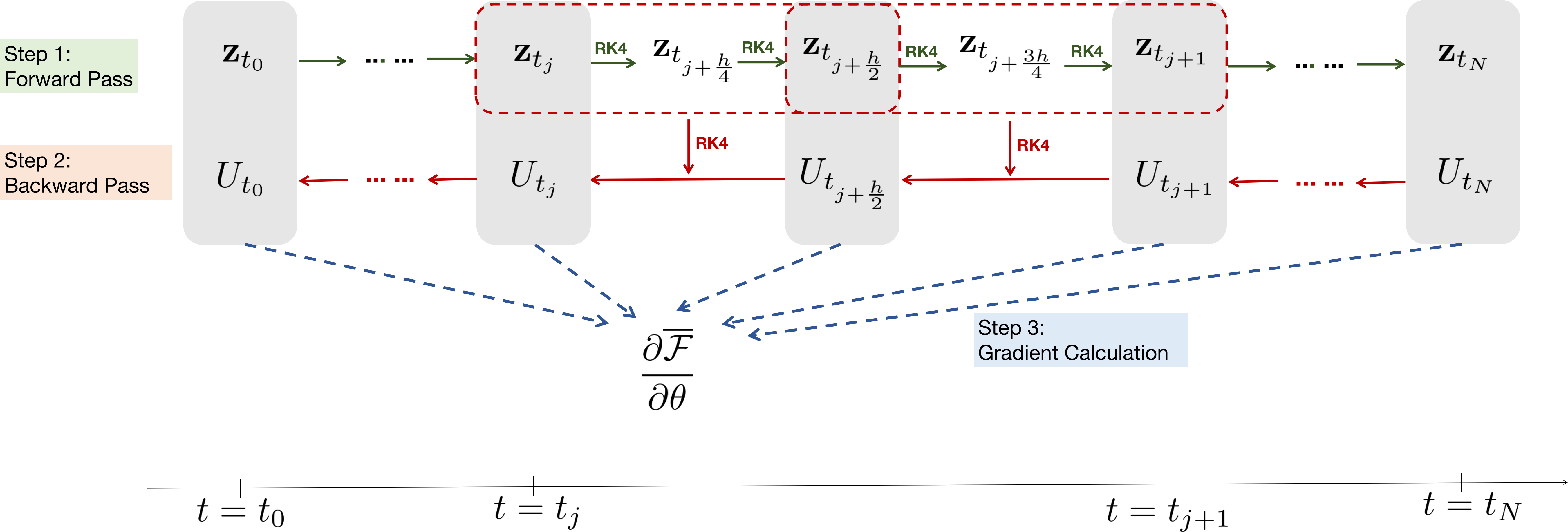}
\caption{The illustration of the adjoint state method.}
\end{center}
\label{flowchart_adjoint}
\end{figure}

The detailed proof of the above proposition is shown in Supplementary Materials. 
For the adjoint state method, the computation of the gradients follows three steps: forward pass, backward pass and gradient calculation, which are different from the back propagation method. Firstly, the forward pass is realized by using the fourth-order explicit Runge-Kutta scheme only to solve \eqref{forwd1} to get the activation throughout time $\mathbf{z}(t)$. Then, the terminal value condition of the adjoint ODE, $U_{t_N}=U(t=1)$, can be further computed by plugging in the values for $\mathbf{z}_{t_N}=\mathbf{z}(t=1)$. During the backward pass, we need to solve the adjoint ODE \eqref{adjoint_eq} to compute $U(t)$ by using all activations $\mathbf{z}(t)$ and still adopting the fourth-order explicit Runge-Kutta scheme. Finally, the gradient with respect to the network parameters $\frac{\partial \overline{\mathcal{F}}}{\partial \theta}$ can be computed by using $\mathbf{z}(t)$ and $U(t)$ from \eqref{adjoint_grad}. The computational flow chart for calculating the gradients using the adjoint state method is shown in Figure \ref{flowchart_adjoint}. It is worth to mention that in order to reduce the memory storage for storing intermediate activations $\mathbf{z}(t)$, only the activation of the last state $\mathbf{z}(1)$ need to be saved in the forward pass, while the intermediate states $\mathbf{z}(t)$ will be recomputed by solving the forward ODE \eqref{forwd1} backward in time during the backward pass.

\subsection{Neural Network Architecture}
\label{NNArc}
In order to solve high-dimensional OT problem, we parameterize the velocity field by a neural network, which can naturally lead to a mesh-free scheme by combining Lagrangian method and Monte Carlo integration.
We now introduce our neural network parameterization of the velocity field.
In the following, we denote our model
as 
 \begin{align*}
 \boldsymbol{v}(x,t;\theta) = \sum_{i=0}^M\boldsymbol{v}_i(x;\theta)\phi_i(t),
 \end{align*}
 where $(x,t) \in \mathbb{R}^{d+1}$ is the input feature, $\theta$ is the network parameters and $\phi_i(t)$ is chosen as the nodal basis function in the time direction, i.e.,
\begin{align*}
\phi_i(t)=
\begin{cases}
0,&t\leq t_{i-1},\\
\frac{t-t_{i-1}}{t_i-t_{i-1}}=M(t-t_{i-1}),&t_{i-1}<t\leq t_i,\\
\frac{t_{i+1}-t}{t_{i+1}-t_{i}}=M(t_{i+1}-t),&t_{i}<t\leq t_{i+1},\\
0,&t> t_{i+1},
\end{cases}
\end{align*}
and $M+1$ denotes the number of the basis function.
When $M$ is fixed, we can choose $N$ (the number of time steps in forward pass) as $kM$, such as $k=2$, so that the forward ODEs \eqref{char} and \eqref{lnrhos} can be solved accurately.
Besides, $\boldsymbol{v}_i(x;\theta)$ is the sum of some two-layer neural networks designed as $\boldsymbol{v}_i(x;\theta)=\sum_{l=1}^L[W_{2i}^{(l)}\sigma(W_{1i}^{(l)}x+b_{1i}^{(l)})+b_{2i}^{(l)}]$, $W_{1i}^{(l)} \in \mathbb{R}^{H \times d}$, $W_{2i}^{(l)} \in \mathbb{R}^{d \times H}$, $b_{1i}^{(l)} \in \mathbb{R}^{H}$, $b_{2i}^{(l)} \in \mathbb{R}^{d}$, the network parameters $\theta = \{W_{1i}^{(l)}, W_{2i}^{(l)}, b_{1i}^{(l)}, b_{2i}^{(l)}\}$, $L$ is the width of the network and $\sigma(x)=\tanh(x)$ is the activation function. 
The advantage of the design of our neural network architecture is to reduce the oscillations in the time direction because we use the piecewise linear function to be the basis function. 

Next, to demonstrate the effective of the above network design, we compare it experimentally with a more natural network structure:
$$
\boldsymbol{v}^*(x,t;\theta)=\sum_{l=1}^L[W_2^{(l)}(t)\sigma(W_1^{(l)}(t)x + b_1^{(l)}(t))+b_2^{(l)}(t)]
$$
since $W_1^{(l)}(t) \in \mathbb{R}^{H\times d}, W_2^{(l)}(t)\in \mathbb{R}^{d\times H}, b_1^{(l)}(t)\in \mathbb{R}^H$ and $b_2^{(l)}(t)\in \mathbb{R}^d$ are time dependent and parameterized directly by the nodal basis functions, i.e.
\begin{align*}
W_e^{(l)}(t) = \sum_{i=0}^M W_{ei}^{(l)} \phi_i(t), ~~~~
b_e^{(l)}(t) = \sum_{i=0}^M b_{ei}^{(l)} \phi_i(t) ~~\text{for } ~~e=1,2.
\end{align*}

\begin{figure}[H]

\begin{minipage}[c]{.05\textwidth}
$\boldsymbol{v}$
\end{minipage}
\begin{minipage}[c]{1.50\textwidth}
\includegraphics[width=4cm]{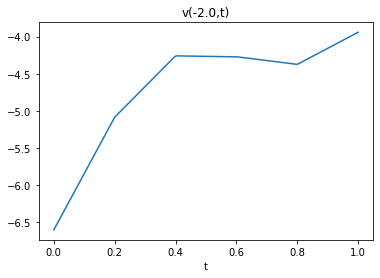}
\includegraphics[width=4cm]{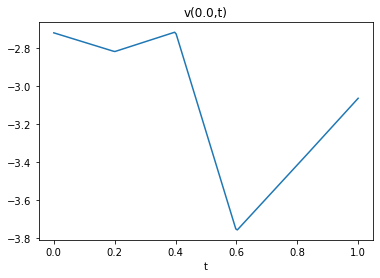}
\includegraphics[width=4cm]{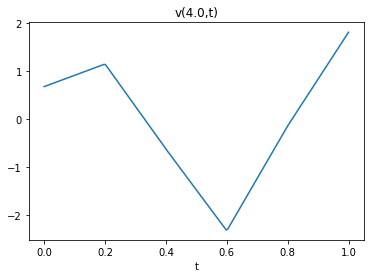}
\end{minipage}

\begin{minipage}[c]{.05\textwidth}
$\boldsymbol{v}^*$
\end{minipage}
\begin{minipage}[c]{1.50\textwidth}
\includegraphics[width=4cm]{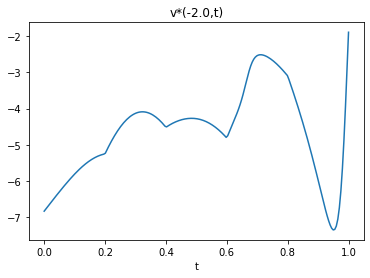}
\includegraphics[width=4cm]{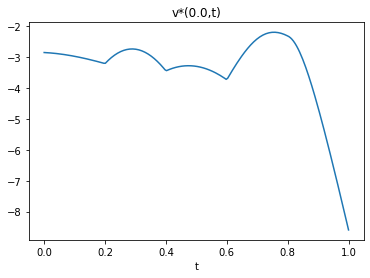}
\includegraphics[width=4cm]{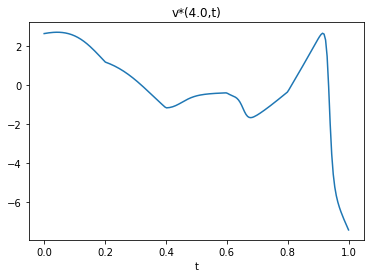}
\end{minipage}
\caption{The comparison of network structures $\boldsymbol{v}$ and $\boldsymbol{v}^*$.}
\label{v}
\end{figure}

We have found that these two velocity fields $\boldsymbol{v}(x,t;\theta)$ and $\boldsymbol{v}^*(x,t;\theta)$ have different network structures, however, the same number of parameters, i.e., $L(M+1)(2dH+H+d)$. Therefore, in order to compare these two neural networks' architecture fairly and intuitively, we give an example in the 1-dimensional case ($d=1$) for Gaussian problem. The initial density $\rho_0$ is a Gaussian with means $0$ and variance $0.3$, while the target density $\rho_1$ is a Gaussian with means $-4$ and variance $1.0$.
The parameters of the network structure are chosen as $L = 2, M = 5$ and $H=10$ for both velocity fields. 
We train our networks using 1000 iterations of Adam with a learning rate of $0.01$. 
We visualize the trained velocity field $\boldsymbol{v}(x,t;\theta)$ and $\boldsymbol{v}^*(x,t;\theta)$ at $x=-2,0,4$ respectively in Figure \ref{v}. 
It can be observed that, on the one hand, the velocity field $\boldsymbol{v}(x,t;\theta)$ is piecewise linear with respect to time, while $\boldsymbol{v}^*(x,t;\theta)$ is nonlinear with respect to time, on the other hand, the oscillation of velocity field $\boldsymbol{v}$ in the time direction is obviously smaller than that of the velocity field $\boldsymbol{v}^*$, both in terms of frequency and amplitude. 
The frequency of $\boldsymbol{v}$ is exactly 5, which is actually equal to $M$ and clearly smaller than the frequency of the velocity field $\boldsymbol{v}^*$. 
In addition, the maximum oscillation amplitude of $\boldsymbol{v}$ is about 4.5, while the maximum oscillation amplitude of $\boldsymbol{v}^*$ is about 11. Since we need to solve ODEs in the time direction, it is difficult to control the numerical accuracy if the velocity field oscillates too much, so the velocity field $\boldsymbol{v}$ is more beneficial to the control accuracy.

\subsection{Crowd Motion}
Our method can also be easily applied to the crowd motion problem \cite{Ruthotto9183}. In contrast to the OT problem, the crowd motion problem not only needs to move the  initial  density $\rho_{0}$ approximately to the  target density $\rho_{1}$, but also  expects to avoid some obstacles in the dynamic movement. The obstacles can be expressed by introducing an added term to the loss function, i.e. the preference term:
\begin{align}\label{prefernceterm}
\mathcal{Q}(\rho) =\lambda_P \int_0^1\int_{\mathbb{R}^d} Q(x)\rho(x,t)\mathrm{d}x \mathrm{d}t,
\end{align}
where $\lambda_P$ is the penalty parameter, and $Q:\mathbb{R}^d  \rightarrow \mathbb{R}$ models the spatial preferences of agents, i.e. the larger $Q(x)$ is, the closer the position of $x$ is to the obstacles. We can get the following formulation:
\begin{align*}
    \min_{(\boldsymbol{v},\rho) \in \mathcal{C}(\rho_0,\rho_1)}\mathcal{F}_{cm}(\boldsymbol{v},\rho) := \mathcal{E}(\boldsymbol{v},\rho)  + \mathcal{Q}(\rho),
\end{align*}
where $\mathcal{E}(\boldsymbol{v},\rho)$ and $\mathcal{C}(\rho_0,\rho_1)$ are what we defined in Section \ref{PDOT}:
\begin{align*}
    \mathcal{E}(\rho,\boldsymbol{v}) = \int_0^1 \int_{\Omega}& |\boldsymbol{v}(x,t)|^2\rho(x,t)\mathrm{d}x \mathrm{d}t,\\
    \mathcal{C}(\rho_0,\rho_1)=\{(\rho, \boldsymbol{v}): \partial_t \rho &+\nabla\cdot (\rho\boldsymbol{v})=0, \rho(\cdot,0)=\rho_0,\rho(\cdot,1)=\rho_1\}.
\end{align*}
The method to solve the crowd motion problem is almost the same as the OT problem. Firstly, as in Section \ref{Se:RoDOT}, we relax the terminal density as an implicit condition penalized by using KL divergence term $\mathcal{P}$ to get the following problem:
\begin{align}\label{rcrm}
    \min_{(\boldsymbol{v},\rho) \in \mathcal{C}(\rho_0)} \mathcal{E}(\boldsymbol{v},\rho)  + \mathcal{Q}(\rho) +\mathcal{P}(\rho),
\end{align}
where 
\begin{align*}
    \mathcal{P}(\rho)& = \lambda \int_{\Omega}\rho(x,1)\log \frac{\rho(x,1)}{\rho_1(x)}
\mathrm{d}x,\\
    \mathcal{C}(\rho_0)=\{(&\rho, \boldsymbol{v}): \partial_t \rho+\nabla\cdot (\rho \boldsymbol{v})=0, \rho(\cdot,0)=\rho_0\}.
\end{align*}
Then we adopt the Lagrangian discretization method presented in Section \ref{Se:RoDOT} to get the semi-discrete formulation of \eqref{rcrm}:
\begin{align}\label{semcd}
    \min_{\boldsymbol{v} }\overline{\mathcal{E}} (\boldsymbol{v},\rho)  + 
 \overline{\mathcal{P}} (\rho) + \overline{\mathcal{Q}} (\rho)
\end{align}
s.t.
\begin{align}
 &\frac{\mathrm{d}}{\mathrm{d}t}\mathbf{z}(x_i,t)=\boldsymbol{v}(\mathbf{z}(x_i,t),t),~\mathbf{z}(x_i,0)=x_i, ~x_i \sim \rho_0(x_{i}), \label{cwmforwd1}\\
 &\frac{\mathrm{d}}{\mathrm{d}t}\ln\rho(\mathbf{z}(x_i,t),t)=-\nabla \cdot \boldsymbol{v}(\mathbf{z}(x_i,t),t),~\ln\rho(\mathbf{z}(x_i,0),0)=\ln\rho_0(x_i),\label{cwmforwd2}
 \end{align}
where $\overline{\mathcal{E}}, \overline{\mathcal{P}}$ are defined in \eqref{semdisE},\eqref{semdisP} respectively, 
\begin{align*}
\overline{\mathcal{E}} (\boldsymbol{v},\rho)&:= \int_0^1 \sum_{i=1}^r \frac{1}{r}|\boldsymbol{v}(\mathbf{z}(x_i,t),t)|^2 \mathrm{d}t, \\
\overline{\mathcal{P}} (\rho)&:= {\lambda}\sum_{i=1}^r \frac{1}{r} \log \frac{\rho(\mathbf{z}(x_i,1),1)}{\rho_1(\mathbf{z}(x_i,1))},
\end{align*}
and $\overline{\mathcal{Q}}$ is the semi-discrete version of $\mathcal{Q}$:
\begin{align*}
    \overline{\mathcal{Q}} (\rho)&:= \int_0^1 \sum_{i=1}^r \frac{1}{r}Q(\mathbf{z}(x_i,t)) \mathrm{d}t.
\end{align*}
The velocity field $\boldsymbol{v}$ is parameterized by neural network $\boldsymbol{v}(x,t;\theta)$ discussed in Section \ref{NNArc}. We discrete \eqref{semcd} by the composite Simpson formula and we use the fourth-order explicit Runge-Kutta scheme to solve ODE constrains \eqref{cwmforwd1} and \eqref{cwmforwd2}.
So that back propagation can be naturally used to calculate the gradient of objective function with respect to network parameters in crowd motion problems. 

\section{Numerical Experiments}\label{Experimentresult}
In this section, we first compare the performance of the adjoint state method with the back propagation method for Gaussian examples (i.e. the initial and target densities are both Gaussian) . 
Next, we demonstrate the performance of the proposed model and its OT results through some numerical experiments. We compare it with several representative and related numerical methods: the Sinkhorn method \cite{cuturi2013sinkhorn} and MFGnet \cite{Ruthotto9183}. We test the effectiveness of our proposed method in high dimensions on some synthetic test problems, including the Gaussian examples and crowd motion examples to perform quantitative and qualitative comparisons, respectively. 
All the experiments of our proposed method are implemented in PyTorch on a NVIDIA Tesla V100 GPU with 32GB memory.

\subsection{Compare the Adjoint State Method with the Back Propagation}\label{AppCwBP}
We consider three typical Gaussian examples with different means and covariance matrices shown in list below. Here $\rho_G(\cdot, \mathbf{m}, \mathbf{\Sigma})$ is the probability density function of a $d$-variate Gaussian with mean $\mathbf{m} \in \mathbb{R}^d$ and covariance matrix $\mathbf{\Sigma} \in \mathbb{R}^{d \times d}$. Since the initial density $\rho_0$ and the target density $\rho_1$ are both Gaussian functions, the ground truth of the transport costs for the OT problem can be exactly known \cite{COTFNT}.\\
\begin{itemize}\label{list1}
    \item Test 1: $\rho_0(x)=\rho_G(x,\mathbf{0},\mathbf{I})$, $\rho_1(x)=\rho_G(x,-4 \cdot \mathbf{e}_1, \mathbf{I})$
    \item Test 2: $\rho_0(x)=\rho_G(x,\mathbf{0},0.3 \cdot \mathbf{I})$, $\rho_1(x)=\rho_G(x,-4 \cdot \mathbf{e}_1, \mathbf{I})$
    \item Test 3: $\rho_0(x)=\rho_G(x,-4 \cdot \mathbf{e}_1 -4 \cdot \mathbf{e}_2,\mathbf{I})$, $\rho_1(x)=\rho_G(x,4 \cdot \mathbf{e}_1 + 4 \cdot \mathbf{e}_2, \mathbf{I})$\\
\end{itemize}

In this section, we compare the adjoint state method with the back propagation method for Gaussian examples. For both methods, the fourth-order explicit Runge-Kutta scheme with a fixed step size is used to solve ODEs and the Simpson formula is adopted for integration in time direction. Besides, for fair comparison, the adjoint state method and the back propagation method use the same structure of the neural network, the same number of time steps and the same learning rate for all the experiments.
\begin{figure}[H]
\begin{center}
\includegraphics[width=2.5cm]{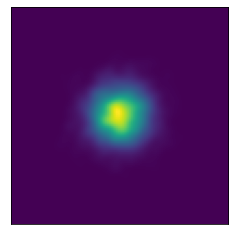}
\includegraphics[width=2.5cm]{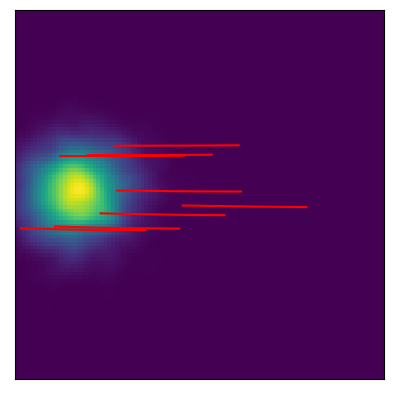}
\includegraphics[width=2.5cm]{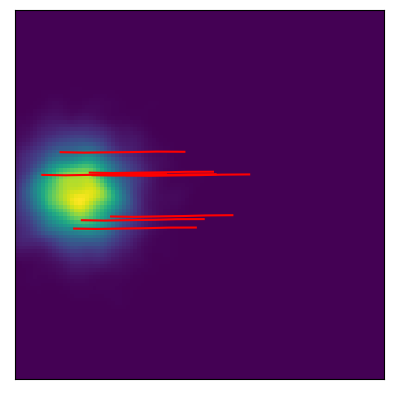}
\includegraphics[width=2.5cm]{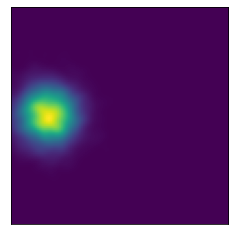}\\
\vspace{5pt}

\includegraphics[width=2.5cm]{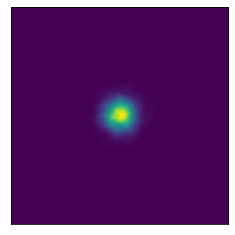}
\includegraphics[width=2.5cm]{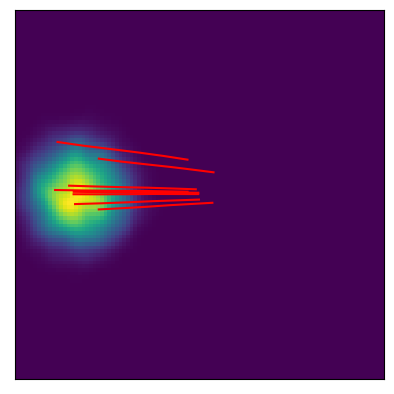}
\includegraphics[width=2.5cm]{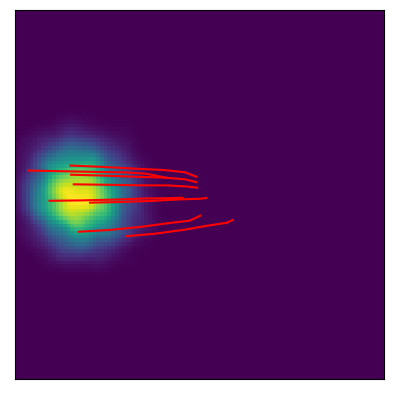}
\includegraphics[width=2.5cm]{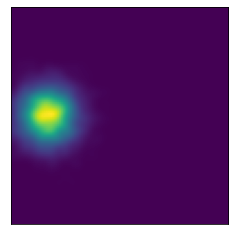}\\

\subfigure[$\rho_0(x_0)$]{\includegraphics[width=2.5cm]{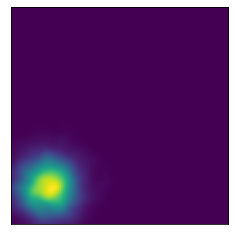}}
\subfigure[BP]{\includegraphics[width=2.5cm]{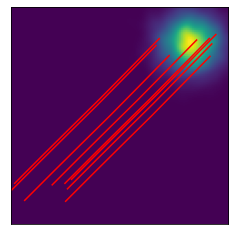}}
\subfigure[Adjoint]{\includegraphics[width=2.5cm]{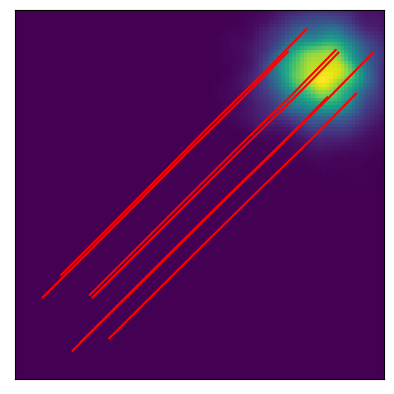}}
\subfigure[$\rho_1(x_1)$]{\includegraphics[width=2.5cm]{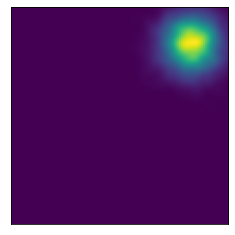}}\\
\vspace{5pt}
\caption{BP v.s. adjoint state method.}
\end{center}
\label{compare}
\end{figure}
In Figure \ref{compare}, we show the experiments results of the back propagation method (b) and the adjoint state method (c) in dimension 2. 
It can be observed that the push forward of $\rho_0$ are very similar to the target density $\rho_1$ for both compared approaches, however, the characteristics are not straight enough for the adjoint state method, especially for Test 2. 
\begin{table}[H]
\newcommand{\tabincell}[2]{\begin{tabular}{@{}#1@{}}#2\end{tabular}} 
\centering
\caption{The comparison of BP and ajoint state method in dimension 2.}
\begin{tabular}{|c| c| c| c|}
\hline
Transport costs &  Test 1 & Test 2 & Test 3\\
\hline
BP & 15.8792 & 16.3305 & 127.7931\\
\hline
Adjoint &15.8792  & 18.1147      &  128.8924 \\
\hline
Ground truth & 16.0000 & 16.4091     & 128.0000\\
\hline
\end{tabular}
\label{adj}
\end{table}

For better comparison, the average values of transport cost obtained with these two methods are also compared in Table \ref{adj}. We can clearly observe that the transport cost estimated by the back propagation is more closer to the ground truth than the results of the adjoint state method. The back propagation method shows better performance than the adjoint state method both quantitatively and qualitatively.
Therefore, in the later experiments, we all use the back propagation method.

\subsection{Gaussian Examples}\label{Gaus_examaple}

In Table \ref{tab:table3}, the average transport costs over five seeds have been calculated and recorded for each method. We can clearly see that the proposed method and MFGnet can lead to better quantitative results than the Sinkhorn method, especially when the dimension is very high. 

\begin{table}[H]
\newcommand{\tabincell}[2]{\begin{tabular}{@{}#1@{}}#2\end{tabular}} 
\centering
\caption{Transport costs of these three instances for different methods in $d=2,10,50$.}
\begin{tabular}{|c|c|c|c|}
\hline
Transport costs & Test 1 & Test 2 & Test 3\\
\hline
\multicolumn{4}{|c|}{$d=2$}\\
\hline
Sinkhorn    & 16.1087  & 16.4795  & 127.3440 \\
\hline
MFGnet & 15.7212 & 16.3645 & 127.9155\\
\hline
Ours(BP) &  15.8792   &  16.3305  & 127.7931 \\ 
\hline
Ground truth   & 16.0000  & 16.4091 & 128.0000\\
\hline
\multicolumn{4}{|c|}{$d=10$}\\
\hline
Sinkhorn          &  18.5652   & 19.4240  & 130.5551  \\
\hline
MFGnet            & 15.6780  & 18.0464 & 127.9691\\
\hline
Ours(BP)         & 16.0131   &  17.9837  & 128.1140 \\ 
\hline
Ground truth     & 16.0000   &  18.0455  & 128.0000 \\
\hline
\multicolumn{4}{|c|}{$d=50$}\\
\hline
Sinkhorn          &  62.1359  &  51.4754   &  174.1247\\
\hline
MFGnet & - & - & -\\
\hline
Ours(BP)         & 16.1800 & 26.0273  & 128.2735 \\ 
\hline
Ground truth     & 16.0000  &  26.2270 & 128.0000\\
\hline
\end{tabular}
\label{tab:table3}
\end{table}

\begin{figure}[H]
\begin{center}
\includegraphics[width=8.0cm]{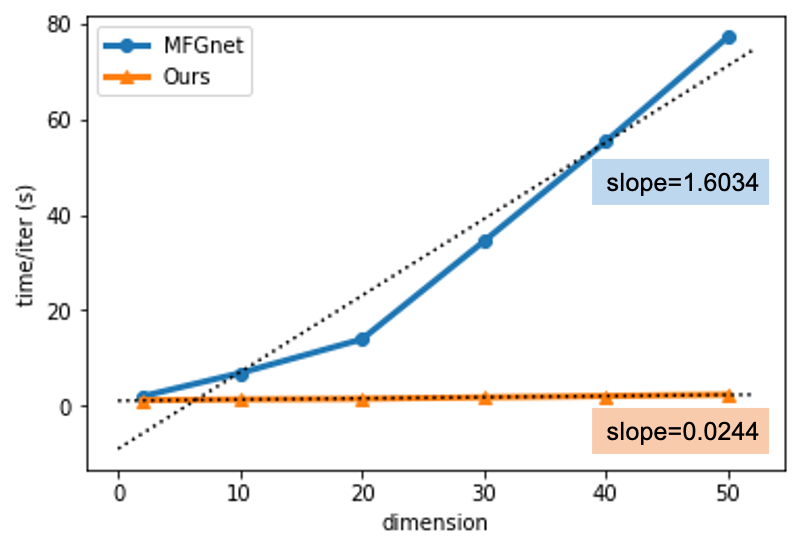}
\vspace{5pt}
\caption{The computational time of each iteration in different dimensions. For MFGnet, the number of training samples are: 2304, 6400, 8464, 10816, 13456 and 16384 at $d=$ 2, 10, 20, 30, 40 and 50, respectively. For the proposed method, 1024 is chosen for all dimensional cases.}
\end{center}
\label{fig:testfig1}
\end{figure}

Note that, for using the Sinkhorn method, firstly, we should discrete the Gaussian distribution by Monte Carlo sampling (see Supplementary Materials for additional experiments of Sinkhorn). Besides, for MFGnet, as the dimension increases, the number of training samples are chosen as 2304, 6400 and 16384 for the 2, 10 and 50 dimensional cases, respectively, as recommended in the literature \cite{Ruthotto9183}. 
Indeed, the running time at dimension 50 is more than 20 hours for each experiment and the BFGS method used in \cite{Ruthotto9183} sometimes fails in the line search, therefore the results are marked with the symbol ``-'' in Table \ref{tab:table3}. 
In addition, we can observe from Figure \ref{fig:testfig1} that the computational time of MFGnet will greatly increase and the slope of growth is about 1.6034. 
For the proposed method, the number of training samples is set as 1024 for all dimensions, and the computational efficiency is only slightly influenced by the increase in dimension (see Figure \ref{fig:testfig1}). The growth slope of the proposed method is only 0.0244, which is much lower than MFGnet. These experiments demonstrate that the proposed method is faster and more efficient than MFGnet and Sinkhorn methods for high-dimensional dynamic OT problems.

\begin{figure}[H]
\begin{center}
\includegraphics[width=2.5cm]{images/test1/test1_d2_push0.png}
\includegraphics[width=2.5cm]{images/test1/test1_2d_characteristic.png}
\includegraphics[width=2.5cm]{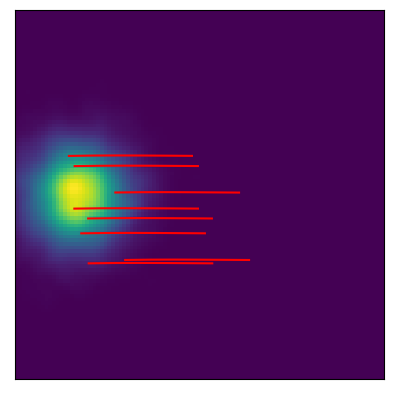}
\includegraphics[width=2.5cm]{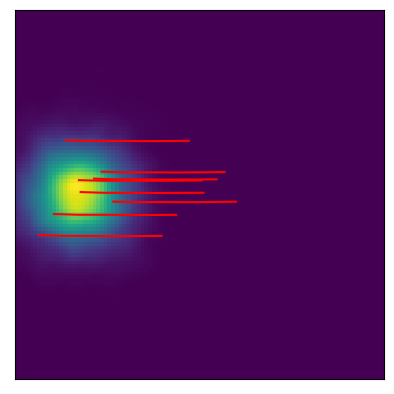}
\includegraphics[width=2.5cm]{images/test1/test1_true.png}\\
\vspace{5pt}

\includegraphics[width=2.5cm]{images/test4/test4_d2_push0.png}
\includegraphics[width=2.5cm]{images/test4/test4_2d_characteristic.png}
\includegraphics[width=2.5cm]{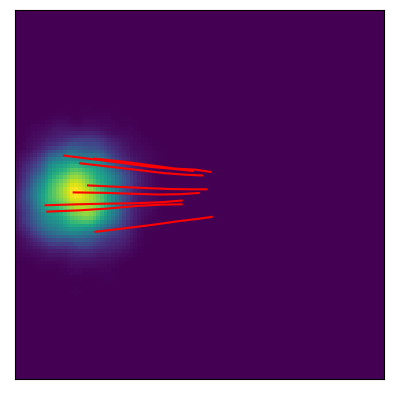}
\includegraphics[width=2.5cm]{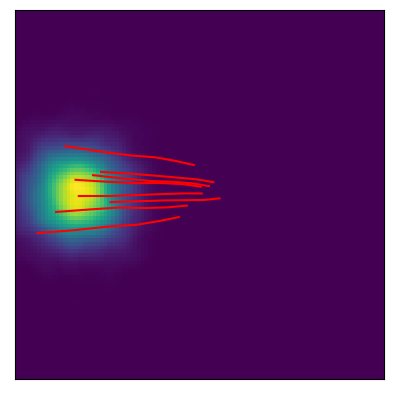}
\includegraphics[width=2.5cm]{images/test4/test4_true.png}\\

\subfigure[$\rho_0(x_0)$]{\includegraphics[width=2.5cm]{images/test2/test2_d2_push0.png}}
\subfigure[$d=2$]{\includegraphics[width=2.5cm]{images/test2/test2_2d_characteristic.png}}
\subfigure[$d=10$]{\includegraphics[width=2.5cm]{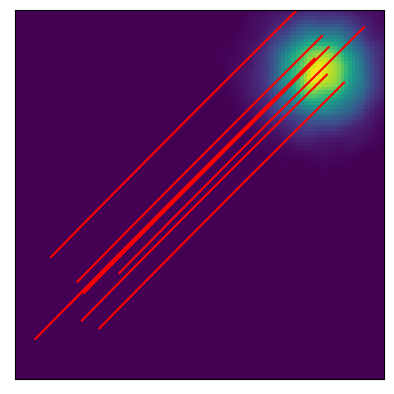}}
\subfigure[$d=50$]{\includegraphics[width=2.5cm]{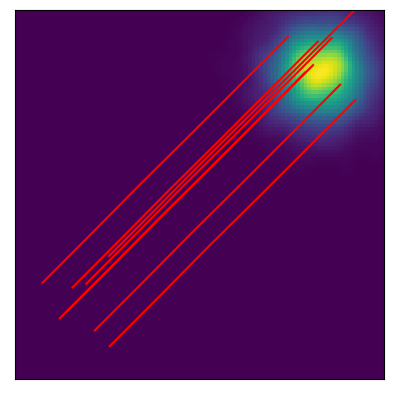}}
\subfigure[$\rho_1(x_1)$]{\includegraphics[width=2.5cm]{images/test2/test2_true.png}}\\

\vspace{5pt}
\caption{The density evolution of Gaussian examples.}
\end{center}
\label{fig:testfig2}
\end{figure}

Next, for better illustration, we also show the density evolution of the proposed method moving from the initial distribution to the target one in Figure \ref{fig:testfig2}. The first column (a) and the last (e) column of Figure \ref{fig:testfig2} show the initial density $\rho_0$ and the target density $\rho_1$, respectively. 
The other columns ((b)-(d)) correspond to the results of the push forward of the initial density $\rho_0$ in dimension $d=2,10$ and $50$, respectively.
We can clearly see that the target density estimated by our trained network shown in (b)-(d) are very similar to the target density $\rho_1$, both in the means and covariance matrices, for all these high-dimensional instances. Note that for $d>2$, we just show slices along the first two coordinate directions. Besides, the red lines represent the characteristics (i.e. the learned trajectories) starting from randomly sampled points according to the initial density $\rho_0$.
It can be observed that the characteristics are almost straight lines, which conforms the density movement is reasonable and the training is successful.

\subsection{Crowd Motion Examples}

In this section, we mainly consider the crowd motion problem which is a generalization of the optimal transport problem. In particular, when $\lambda_P=0$ in \eqref{prefernceterm}, the crowd motion problem will degenerate into an OT problem.

In list below, we list two representative tests with different choices of initial density $\rho_0$, target density $\rho_1$ and preference function $Q(x)$ is Gaussian or Gaussian mixture, corresponding to one obstacle (Test 4) and two obstacles (Test 5) respectively. Note that for $d>2$,  we just use the first two components of $x$ to evaluate the preference function $Q(x)$.\\

\begin{itemize}\label{lst:crowdmotion}
      \item Test4: $\rho_0(x)=\rho_G(x,-3 \cdot \mathbf{e}_1, 0.3 \cdot \mathbf{I})$, $\rho_1(x)=\rho_G(x,3 \cdot \mathbf{e}_1, 0.3 \cdot \mathbf{I})$,\\
      $~~~~~~~~~~Q(x) = \rho_G(x,\mathbf{0},diag(0.5,1))$
      \item Test5: $\rho_0(x)=\rho_G(x,-4 \cdot \mathbf{e}_1, 0.3 \cdot \mathbf{I})$, $\rho_1(x)=\rho_G(x,4 \cdot \mathbf{e}_1, 0.3 \cdot \mathbf{I})$,\\
 $~~~~~~Q(x) = \frac{1}{2}\rho_G(x,-2 \cdot \mathbf{e}_2,diag(0.1,1))+\frac{1}{2}\rho_G(x,2\cdot \mathbf{e}_2,diag(0.1,1))$\\
\end{itemize}

\begin{figure}[h]
\begin{minipage}[c]{.18\textwidth}
$\rho_0(x_0)$
\end{minipage}
\begin{minipage}[c]{.80\textwidth}
\includegraphics[width=2.5cm]{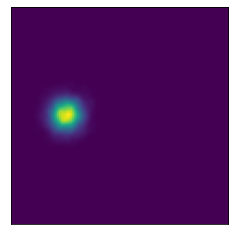}
\includegraphics[width=2.5cm]{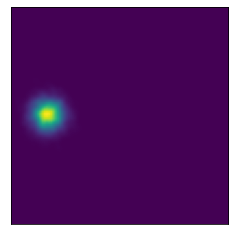}
\includegraphics[width=2.5cm]{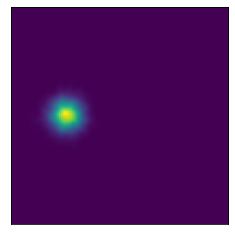}
\includegraphics[width=2.5cm]{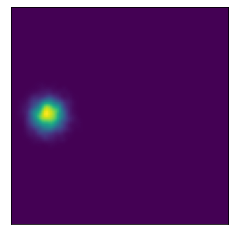}
\end{minipage}

\begin{minipage}[c]{.18\textwidth}
$\tilde{\rho}_{1/5}(\tilde{x}_{1/5})$
\end{minipage}
\begin{minipage}[c]{.80\textwidth}
\includegraphics[width=2.5cm]{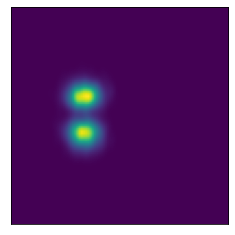}
\includegraphics[width=2.5cm]{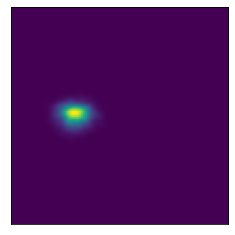}
\includegraphics[width=2.5cm]{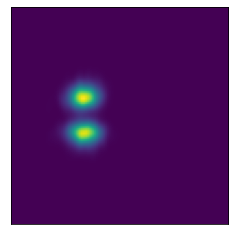}
\includegraphics[width=2.5cm]{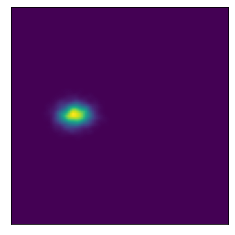}
\end{minipage}

\begin{minipage}[c]{.18\textwidth}
$\tilde{\rho}_{1/2}(\tilde{x}_{1/2})$
\end{minipage}
\begin{minipage}[c]{.80\textwidth}
\includegraphics[width=2.5cm]{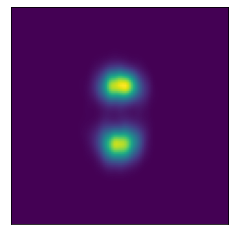}
\includegraphics[width=2.5cm]{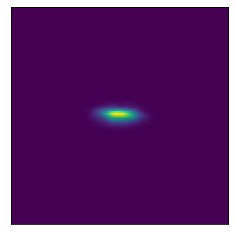}
\includegraphics[width=2.5cm]{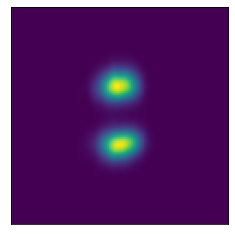}
\includegraphics[width=2.5cm]{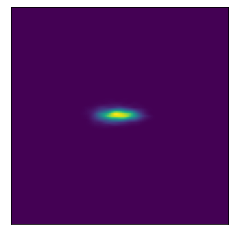}
\end{minipage}

\begin{minipage}[c]{.18\textwidth}
$\tilde{\rho}_{4/5}(\tilde{x}_{4/5})$
\end{minipage}
\begin{minipage}[c]{.80\textwidth}
\includegraphics[width=2.5cm]{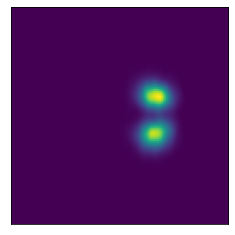}
\includegraphics[width=2.5cm]{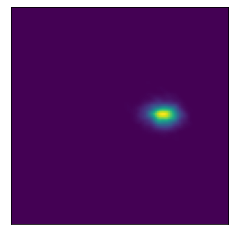}
\includegraphics[width=2.5cm]{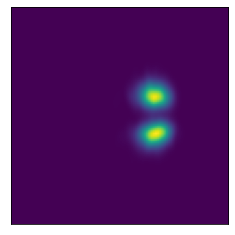}
\includegraphics[width=2.5cm]{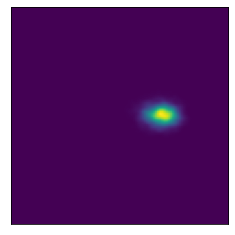}
\end{minipage}

\begin{minipage}[c]{.18\textwidth}
$\tilde{\rho}_1(\tilde{x}_1)$
\end{minipage}
\begin{minipage}[c]{.80\textwidth}
\includegraphics[width=2.5cm]{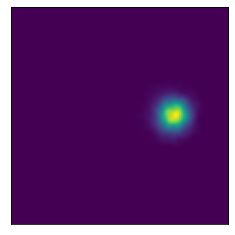}
\includegraphics[width=2.5cm]{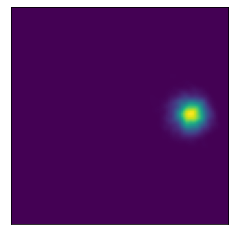}
\includegraphics[width=2.5cm]{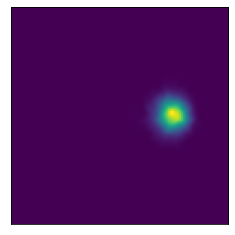}
\includegraphics[width=2.5cm]{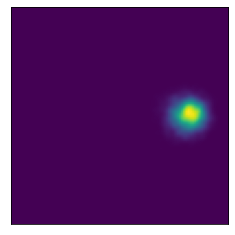}
\end{minipage}

\begin{minipage}[c]{.18\textwidth}
$\rho_{1}(x_{1})$
\end{minipage}
\begin{minipage}[c]{.80\textwidth}
\includegraphics[width=2.5cm]{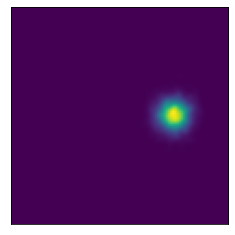}
\includegraphics[width=2.5cm]{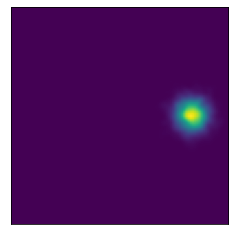}
\includegraphics[width=2.5cm]{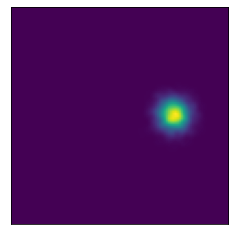}
\includegraphics[width=2.5cm]{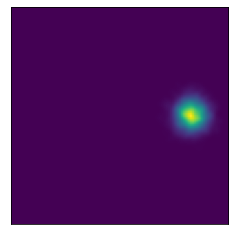}
\end{minipage}

\begin{minipage}[c]{.18\textwidth}
obstacles \\
+\\
characteristics 
\end{minipage}
\begin{minipage}[c]{.80\textwidth}
\includegraphics[width=2.5cm]{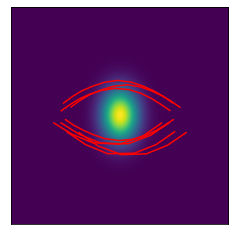}
\includegraphics[width=2.5cm]{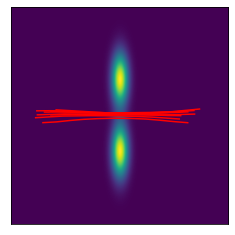}
\includegraphics[width=2.5cm]{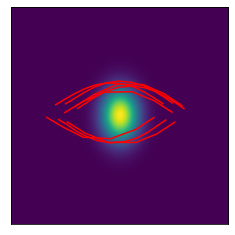}
\includegraphics[width=2.5cm]{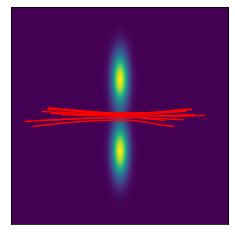}
\end{minipage}
\\
\vspace{5pt}
\caption{Illustration of the crowd motion problem at $d=2$ (left two columns) and $d=5$ (right two columns).}
\label{crowdmotion1}
\end{figure}

We set the main parameters as
$\lambda = 10$, $\lambda_{P} = 500$ and $\alpha = 0$ and we focus on the case $d=2$ (left two columns) and $d=5$ (right two columns) in Figure \ref{crowdmotion1}. The last row shows the obstacles in the center of the domain and the red lines present the learned characteristics. The remaining rows in Figure \ref{crowdmotion1} show the density evolution at intermediate time $t=\frac{1}{5}, \frac{1}{2}, \frac{4}{5}, 1$ from the initial density $\rho_0$ to the target density $\rho_1$. We can clearly observe that the push forward of $\rho_0$ (i.e. $\tilde{\rho}_1(\tilde{x}_1)$) is similar to the target density $\rho_1$ and the characteristics are curved, not straight, to avoid the regions where these obstacles are located.

\begin{figure}[h]
\begin{minipage}[c]{.18\textwidth}
$\rho_0(x_0)$
\end{minipage}
\begin{minipage}[c]{.80\textwidth}
\includegraphics[width=2.5cm]{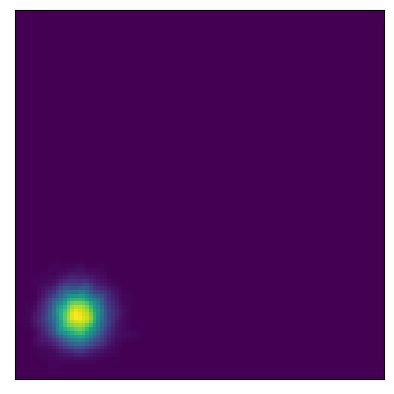}
\includegraphics[width=2.5cm]{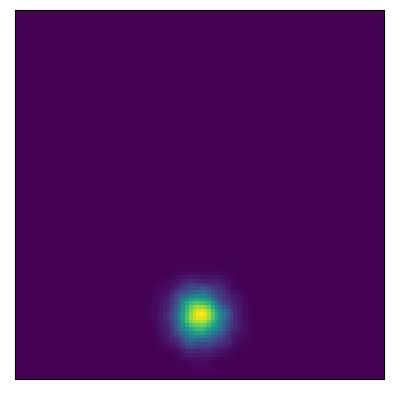}
\includegraphics[width=2.5cm]{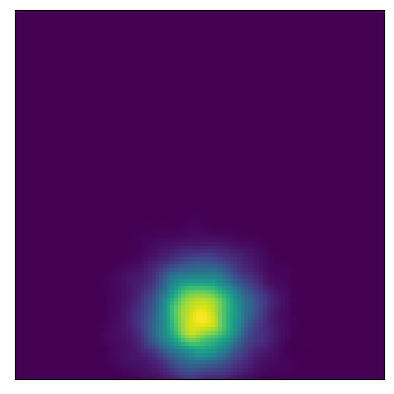}
\includegraphics[width=2.5cm]{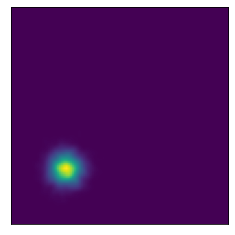}
\end{minipage}

\begin{minipage}[c]{.18\textwidth}
$\tilde{\rho}_{1/5}(\tilde{x}_{1/5})$
\end{minipage}
\begin{minipage}[c]{.80\textwidth}
\includegraphics[width=2.5cm]{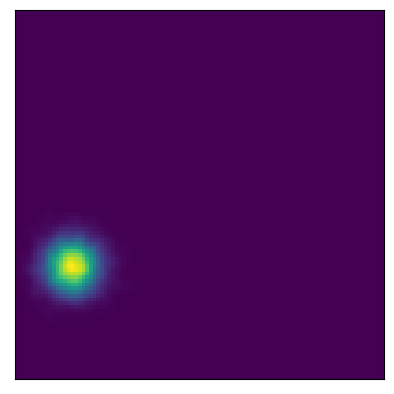}
\includegraphics[width=2.5cm]{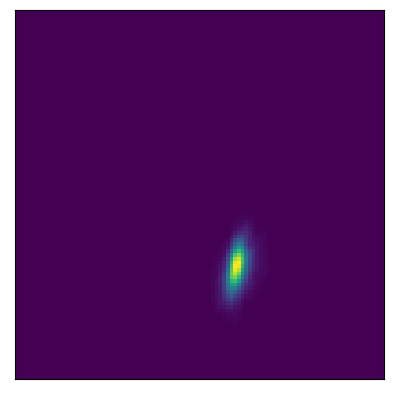}
\includegraphics[width=2.5cm]{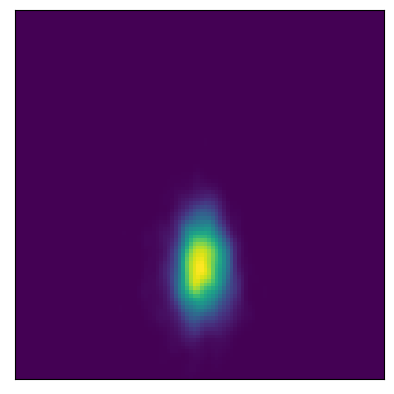}
\includegraphics[width=2.5cm]{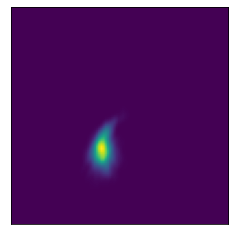}
\end{minipage}

\begin{minipage}[c]{.18\textwidth}
$\tilde{\rho}_{1/2}(\tilde{x}_{1/2})$
\end{minipage}
\begin{minipage}[c]{.80\textwidth}
\includegraphics[width=2.5cm]{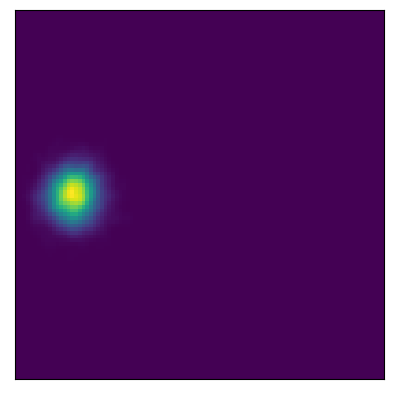}
\includegraphics[width=2.5cm]{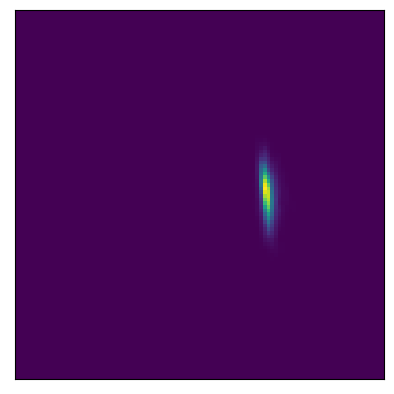}
\includegraphics[width=2.5cm]{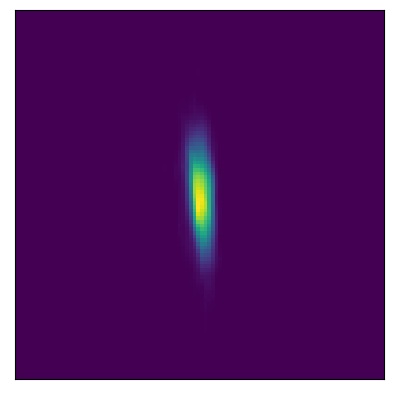}
\includegraphics[width=2.5cm]{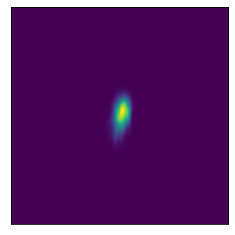}
\end{minipage}

\begin{minipage}[c]{.18\textwidth}
$\tilde{\rho}_{4/5}(\tilde{x}_{4/5})$
\end{minipage}
\begin{minipage}[c]{.80\textwidth}
\includegraphics[width=2.5cm]{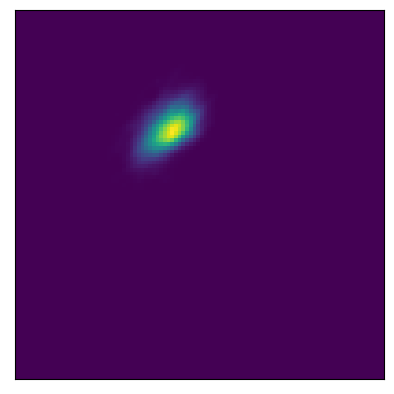}
\includegraphics[width=2.5cm]{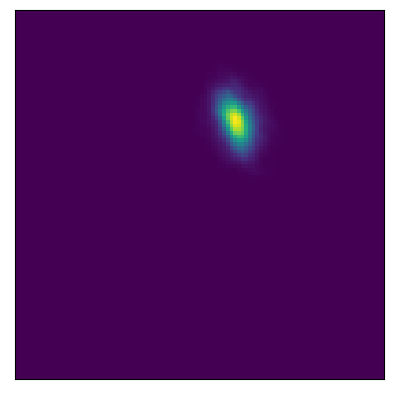}
\includegraphics[width=2.5cm]{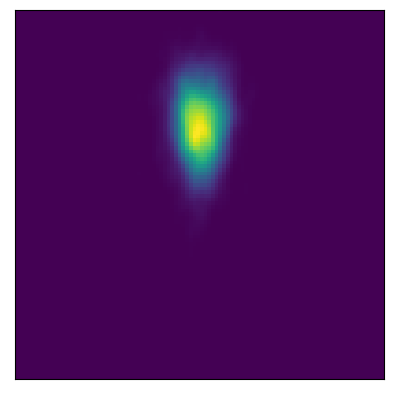}
\includegraphics[width=2.5cm]{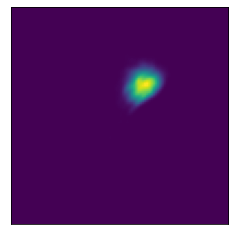}
\end{minipage}

\begin{minipage}[c]{.18\textwidth}
$\tilde{\rho}_1(\tilde{x}_1)$
\end{minipage}
\begin{minipage}[c]{.80\textwidth}
\includegraphics[width=2.5cm]{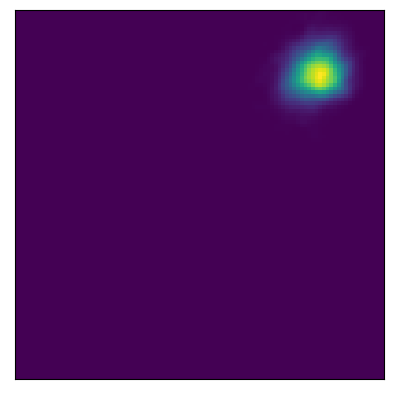}
\includegraphics[width=2.5cm]{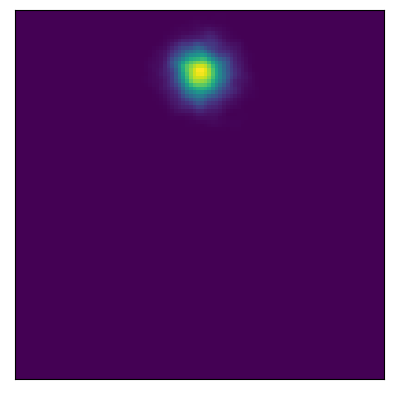}
\includegraphics[width=2.5cm]{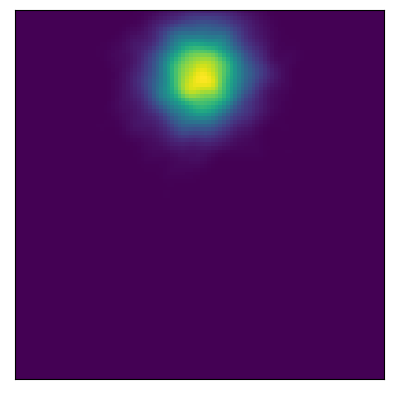}
\includegraphics[width=2.5cm]{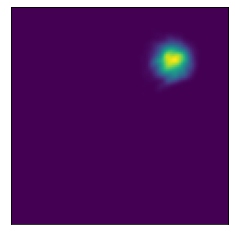}
\end{minipage}

\begin{minipage}[c]{.18\textwidth}
$\rho_{1}(x_{1})$
\end{minipage}
\begin{minipage}[c]{.80\textwidth}
\includegraphics[width=2.5cm]{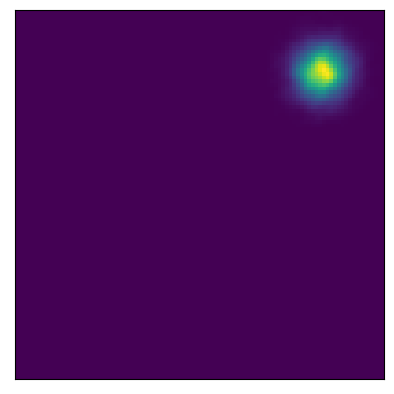}
\includegraphics[width=2.5cm]{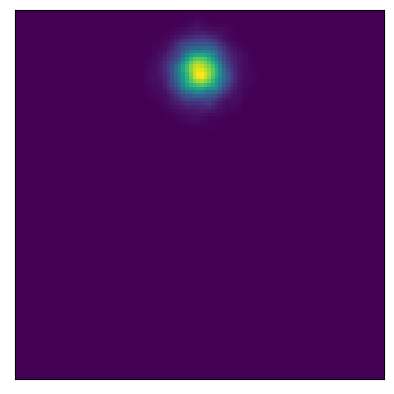}
\includegraphics[width=2.5cm]{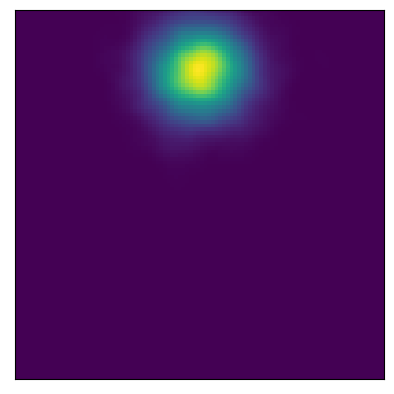}
\includegraphics[width=2.5cm]{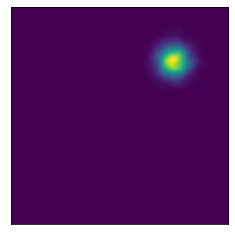}
\end{minipage}

\begin{minipage}[c]{.18\textwidth}
obstacles \\
+\\
characteristics 
\end{minipage}
\begin{minipage}[c]{.80\textwidth}
\includegraphics[width=2.5cm]{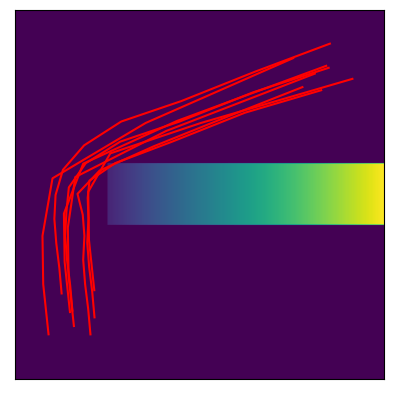}
\includegraphics[width=2.5cm]{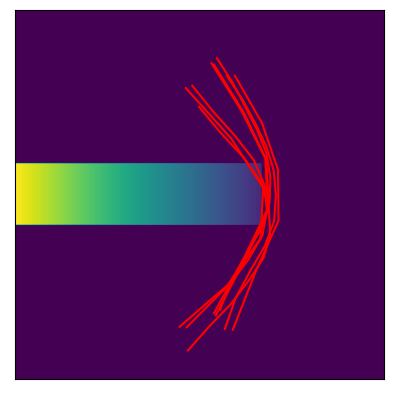}
\includegraphics[width=2.5cm]{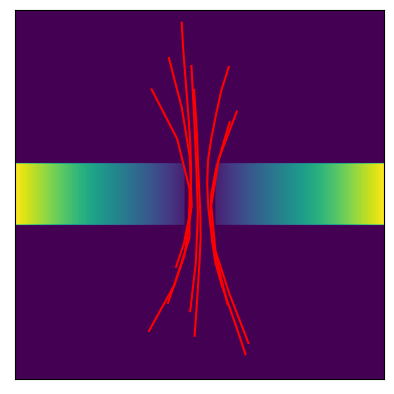}
\includegraphics[width=2.5cm]{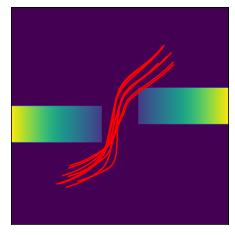}
\end{minipage}
\\
\vspace{5pt}
\caption{Illustration of the crowd motion problem at $d=2$ for maze examples.}
\label{maze}
\end{figure}

Our second crowd motion experiment simulates some maze examples in Figure \ref{maze} and obstacles are represented by an indicator function where the regions are regular rectangles. We show the crowd motion results of two-dimensional case with different obstacles. In order to avoid obstacles successfully, firstly, we adopt the same method as Gaussian function to blur rectangular obstacles properly shown in the last row of Figure \ref{maze}.
Different colors indicate the strength of the obstruction in different parts of the obstacles, that is, the darker the color, the smaller the obstruction. The density evolution and characteristics show that the mass can circumvent the obstacles very well, which demonstrates the success of our algorithm in the crowd motion problem.

\section{Conclusion}\label{ConclusionandFutureWork}

Motivated by the fact that the traditional mesh-based discretization method usually leads to the curse of dimensionality, we propose a novel neural network based method for solving dynamic OT problems effectively in very high-dimensional spaces.
In order to calculate the gradient accurately, we have adopted two different method: the back propagation method and the adjoint state method. The numerical experiments demonstrate that the proposed back propagation method performs well compared to the MFGnet and the Sinkhorn method. Our proposed method is easy to implement and high precision in experiments. 
Generally speaking, for different dimension, our proposed method takes about 1.5s to perform one iteration with our unoptimized PyTorch codes and GPU implementation.
In the future, we will further study how to extend this method to the unnormalized OT problems, which means initial and target densities with different total mass.



\bibliographystyle{siam}
\bibliography{bibtex}

\end{document}